\documentclass[11pt,letterpaper]{article}
\usepackage{emnlp2017}
\usepackage{times}
\usepackage{latexsym}
\usepackage{graphicx}
\usepackage{mathtools}
\usepackage{subcaption}
\usepackage{caption}
\usepackage{xcolor}
\usepackage{multirow}
\usepackage{amssymb}
\usepackage{bm}
\emnlpfinalcopy

\def\emnlppaperid{852}

\title{CharManteau: Character Embedding Models For Portmanteau Creation}

\author{Varun Gangal \thanks{* denotes equal contribution},  Harsh Jhamtani \footnotemark[1], Graham Neubig, Eduard Hovy, Eric Nyberg \\
Language Technologies Institute, \\ Carnegie Mellon University \\
  {\tt \{vgangal,jharsh,gneubig,hovy,ehn\}@cs.cmu.edu} }

\date{}

\begin{document}

\maketitle

\begin{abstract}
Portmanteaus are a word formation phenomenon where two words 
are combined 
to form a new word. We propose character-level neural sequence-to-sequence (S2S) methods for the task of portmanteau generation that are end-to-end-trainable, language independent, and do not explicitly use additional phonetic information. 
We propose a noisy-channel-style model, which allows for the incorporation of unsupervised word lists, improving performance over a standard source-to-target model. This model is made possible by an exhaustive candidate generation strategy specifically enabled by the features of the portmanteau task. Experiments find our approach superior to a state-of-the-art FST-based baseline with respect to ground truth accuracy and human evaluation. 
\end{abstract}

\section{Introduction}
\label{sec:intro}

Portmanteaus (or lexical blends \newcite{algeo1977blends}) are novel words formed from parts of multiple root words in order to refer to a new concept which can't otherwise be expressed concisely. Portmanteaus have become frequent in modern-day social media, news reports and advertising, one popular example being \textit{Brexit} (Britain + Exit). \newcite{washingtonpost}. These are found not only in English but many other languages such as Bahasa Indonesia \newcite{dardjowidjojo1979acronymic}, Modern Hebrew \newcite{bat1996selecting,berman1989role} and Spanish \newcite{pineros2004creation}. Their short length makes them ideal for headlines and brandnames \cite{nyt}.
\begin{figure}
    \centering
 \includegraphics[scale=0.34]{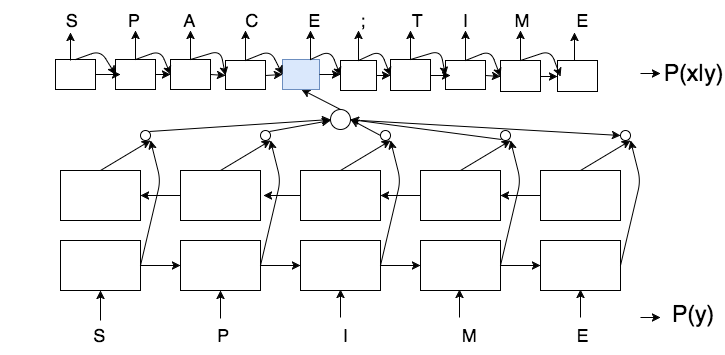}
    \caption{A sketch of our \textsc{Backward}, noisy-channel model. The attentional S2S model with bidirectional encoder gives $P(x|y)$ and next-character model gives $P(y)$, where $y$ (\textit{spime}) is the portmanteau and $\bm{x}=\text{concat}(\bm{x^{(1)}},\text{``;''},\bm{x^{(2)}})$ are the concatenated root words (\textit{space} and \textit{time}).
    }
    \textbf{ \label{fig:arch}}
\end{figure}
Unlike better-defined morphological phenomenon such as inflection and derivation, portmanteau generation is difficult to capture using a set of rules.
For instance, \newcite{shaw2014emergent} state that the composition of the portmanteau from its root words depends on several factors, two important ones being maintaining prosody and retaining character segments from the root words, especially the head.
An existing work by \newcite{deri2015make} aims to solve the problem of predicting portmanteau using a multi-tape FST model, which is data-driven, unlike prior approaches.
Their methods rely on a grapheme to phoneme converter, which takes into account the phonetic features of the language, but may not be available or accurate for non-dictionary words, or low resource languages.

Prior works, such as \newcite{faruqui2016morphological}, have demonstrated the efficacy of neural approaches for morphological tasks such as inflection.
We hypothesize that such neural methods can
(1) provide a simpler and more integrated end-to-end framework than multiple FSTs used in the previous work, and (2) automatically capture features such as phonetic similarity through the use of character embeddings, removing the need for explicit grapheme-to-phoneme prediction.
To test these hypotheses, in this paper, we propose a neural S2S model to predict portmanteaus given the two root words, specifically making 3 major contributions:
\begin{itemize}
\itemsep-0.3em 
\item We propose an S2S model that attends to the two input words to generate portmanteaus, and an additional improvement that leverages noisy-channel-style modelling to incorporate a language model over the vocabulary of words (\S\ref{sec:models}).
\item Instead of using the model to directly predict output character-by-character, we use the features of portmanteaus to exhaustively generate candidates, making scoring using the noisy channel model possible (\S\ref{sec:predictions}).
\item We curate and share a new and larger dataset of 1624 portmanteaus (\S\ref{sec:dataset}). 
\end{itemize}
In experiments (\S\ref{sec:experiments}), our model performs better than the baseline \newcite{deri2015make} on both objective and subjective measures, demonstrating that such methods can be used effectively in a morphological task.

\section{Proposed Models}
\label{sec:models}





This section describes our neural models.

\subsection{Forward Architecture}

Under our first proposed architecture, the input sequence $\bm{x} = \text{concat}(\bm{x}^{(1)},\text{``;''},\bm{x}^{(2)})$, while the output sequence is the portmanteau $\bm{y}$. The model learns the distribution $P(\bm{y}|\bm{x})$. 

The network architecture we use is an attentional S2S model \cite{bahdanau2014neural}. We use a bidirectional encoder, which is known to work well for S2S problems with similar token order, which is true in our case. Let $\overrightarrow{LSTM}$ and $\overleftarrow{LSTM}$ represent the forward and reverse encoder; $e_{enc}()$ and $e_{dec}()$ represent the character embedding functions used by encoder and decoder 
The following equations describe the model: 
\begin{align*}
    h^{\overrightarrow{enc}}_0 &=\overrightarrow{0},  h^{\overleftarrow{enc}}_{|x|} = \overrightarrow{0} \\
    h^{\overrightarrow{enc}}_{t} &= \overrightarrow{LSTM}(h^{enc}_{t-1},{e_{enc}}({x_{t}})) \\
    h^{\overleftarrow{enc}}_{t} &= \overleftarrow{LSTM}(h^{enc}_{t+1},{e_{enc}}({x_{t}})) \\
    h^{enc}_{t} &= h^{\overrightarrow{enc}}_{t} + h^{\overleftarrow{enc}}_{t} \\
    h^{dec}_0 &= h^{enc}_{|x|} \\
    h^{dec}_{t} &= LSTM(h^{dec}_{t-1},[\text{concat}({e_{dec}}({y_{t-1}}),c_{t-1})]) \\
    p_{t}&=\text{softmax}(W_{hs}[\text{concat}(h^{dec}_{t},c_{t})]+b_{s})
\end{align*}
The context vector $c_{t}$ is computed using dot-product attention over encoder states. We choose dot-product attention because it doesn't add extra parameters, which is important in a low-data scenario such as portmanteau generation.

{\small
\begin{align*}
        a^{t}_{i} &= dot(h^{dec}_{t},h^{enc}_{i}) ,         \alpha^{t} = \text{softmax}(a^{t}) \\
        c_{t} &= \sum_{i=1}^{i=|x|} \alpha^{t}_{i} h^{enc}_{i} \\
\end{align*}
}%

In addition to capturing the fact that portmanteaus of two English words typically sound \emph{English-like}, and to compensate for the fact that available portmanteau data will be small, we pre-train the character embeddings on English language words.
 We 
 use character embeddings learnt using 
 an LSTM language model over words in an English dictionary,\footnote{ Specifically in our experiments, 134K words from the CMU dictionary \cite{weide1998cmu}.} where
 each word is a sequence of characters, and the model will predict next character in sequence conditioned on previous characters in the sequence. 

\subsection{Backward Architecture} \label{subsection:back}
The second proposed model uses Bayes's rule to reverse the probabilities
$P(\bm{y}|\bm{x})=\frac{P(\bm{x}|\bm{y})P(\bm{y})}{P(\bm{x})}$ to get $\operatorname*{argmax}_{\bm{y}} P(\bm{y}|\bm{x}) = \operatorname*{argmax}_{\bm{y}} P(\bm{x}|\bm{y})P(\bm{y})$. 
Thus, we have a reverse model of the probability $P(\bm{x}|\bm{y})$ that the given root words were generated from the portmanteau and a character language model model $P(\bm{y})$.
This is a probability distribution over all character sequences $y \in A^{*}$, where $A$ is the alphabet of the language. This way of factorizing the probability is also known as a \textit{noisy channel model}, 
which has recently also been shown to be effective for neural MT (\newcite{hoang2017decoding}, \newcite{yu2016neural}). Such a model offers two advantages
\begin{enumerate}
\itemsep-0.1em
\item The reverse direction model (or alignment model) gives higher probability to those portmanteaus from which one can discern the root words easily, which is one feature of good portmanteaus.
\item 
The character language model $P(\bm{y})$ can be trained on a large vocabulary of words in the language. 
The likelihood of a word $y$ is factorized as $P(y)=\Pi_{i=1}^{i=|y|} P(y_{i}|y_{1}^{i-1})$, where $y_{j}^{i} = y_{i}, y_{i+1} \ldots y_{j}$, and we train a LSTM to maximize this likelihood.
\end{enumerate}

\section{Making Predictions}
\label{sec:predictions}
Given these models, we must make predictions, which we do by two methods 
\begin{description}
\item[Greedy Decoding:]
In most neural sequence-to-sequence models, we perform auto-regressive greedy decoding, selecting the next character greedily based on the probability distribution for the next character at current time step. We refer to this decoding strategy as \textsc{Greedy}.
\item[Exhaustive Generation:]
Many portmanteaus were observed to be concatenation of a prefix of the first word and a suffix of the second. We therefore generate all candidate outputs which follow this rule. Thereafter we score these candidates with the decoder and output the one with the maximum score. We refer to this decoding strategy as \textsc{Score}.
\end{description}

Given that our training data is small in size, we expect ensembling \cite{breiman1996bagging} to help reduce model variance and improve performance. In this paper, we ensemble our models wherever mentioned by training multiple models on 80\% subsamples of the training data, and averaging log probability scores across the ensemble at test-time. 

\section{Dataset}
\label{sec:dataset}
The existing dataset by \newcite{deri2015make} contains 401 portmanteau examples from Wikipedia. We refer to this dataset as $D_{\textit{Wiki}}$. Besides being small for detailed evaluation, $D_{\textit{Wiki}}$ is biased by being from just one source.
We manually collect $D_{\textit{Large}}$, a dataset of 1624 distinct English portmanteaus from following sources:
\begin{itemize}
\itemsep-0.1em 
\item Urban Dictionary\footnote{Not all neologisms are portmanteaus, so we manually choose those which are for our dataset.}
\item Wikipedia
\item Wiktionary 
\item \href{http://tinyurl.com/le4lvyj}{BCU's Neologism Lists} from '94 to '12.
\end{itemize}
Naturally, $D_{\textit{Wiki}} \subset D_{\textit{Large}}$. We define $D_{\textit{Blind}}= D_{\textit{Large}}-D_{\textit{Wiki}}$ as the dataset of 1223 examples not from Wikipedia. 
We observed that 84.7\% of the words in $D_{\textit{Large}}$ can be generated by concatenating prefix of first word with a suffix of the second.

\section{Experiments}
\label{sec:experiments}

In this section, we show results comparing various configurations of our model to the baseline FST model of \newcite{deri2015make} (\textsc{BASELINE}).
Models are evaluated using exact-matches (\textit{Matches}) and average Levenshtein edit-distance (\textit{Distance}) w.r.t ground truth. 
 


\begin{table}
\centering
\scriptsize
\begin{tabular}{|l|l|l|l|l|l|l| }
\hline 
Model & Attn  & Ens & Init & Prediction & Matches & Distance\\ \hline \hline
\textsc{Baseline} & {-} & {-}  &  {-} & {-} & \textbf{45.39\%} & 1.59  \\ \hline
\multirow{10}{*}{\textsc{Forward}} &  $\checkmark$ & $\times$ & $\times$ & \textsc{Greedy} & 22.00\% & 1.98 \\ 
 & $\checkmark$ & $\times$ & $\checkmark$ & \textsc{Greedy} & 28.00\% & 1.90 \\ 
& $\checkmark$ & $\times$ & $\times$ & \textsc{Beam} &  13.25\% & 2.47 \\ 
& $\checkmark$ & $\times$ & $\checkmark$ & \textsc{Beam} &  15.25\% & 2.37 \\  
 & $\checkmark$ &  $\times$ & $\times$ & \textsc{Score} & 30.25\% & 1.64 \\ 
& $\checkmark$ & $\times$ & $\checkmark$ & \textsc{Score} &  32.88\% & 1.53 \\ 
& $\checkmark$ & $\checkmark$ & $\checkmark$ & \textsc{Score} &  42.25\% & 1.33 \\ 
& $\checkmark$ & $\checkmark$ & $\times$ & \textsc{Score} &  41.25\% & 1.34 \\ 

& $\times$ & $\times$ & $\checkmark$ & \textsc{Score} & 6.75\% & 3.78 \\ 
& $\times$ & $\times$ & $\times$ & \textsc{Score} & 6.50\% & 3.76 \\ \hline
\multirow{6}{*}{\textsc{Backward}} & $\checkmark$ & $\times$ & $\times$ & \textsc{Score} & 37.00\% & 1.53 \\ 
 & $\checkmark$ & $\times$ & $\checkmark$ & \textsc{Score} & 42.25\% & 1.35 \\ 
 & $\checkmark$ & $\checkmark$ & $\checkmark$ & \textsc{Score} & \textbf{48.75\%} & \textbf{1.12} \\ 
 & $\checkmark$ & $\checkmark$ & $\times$ & \textsc{Score} &  \textbf{46.50\%} & \textbf{1.24} \\
 & $\times$ & $\times$ & $\checkmark$ & \textsc{Score} & 5.00\% & 3.95 \\ 
 & $\times$ & $\times$ & $\times$ & \textsc{Score} &  4.75\% & 3.98 \\ \hline
\end{tabular}
\caption{10-Fold Cross-Validation results, $D_{\textit{Wiki}}$. \emph{Attn, Ens, Init} denote attention, ensembling, and initializing character embeddings respectively.}
\textbf{\label{tab:knightExp}}
\end{table}



\subsection{Objective Evaluation Results}

In \emph{Experiment 1}, we follow the same setup as \newcite{deri2015make}. $D_{\textit{Wiki}}$ is split into 10 folds. Each fold model uses 8 folds for training, 1 for validation, and 1 for test. The average (10 fold cross-validation style approach) performance metrics on the test fold are then evaluated. \emph{Table \ref{tab:knightExp}} shows the results of \emph{Experiment 1} for various model configurations. We get the \textsc{BASELINE} numbers from \newcite{deri2015make}. Our best model obtains $48.75\%$ \textit{Matches} and $1.12$ \textit{Distance}, compared to $45.39\%$ \textit{Matches} and $1.59$ \textit{Distance} using \textsc{BASELINE}. 


For \emph{Experiment 2}, we seek to compare our best approaches from \emph{Experiment 1} to the \textsc{BASELINE} on a large, held-out dataset. Each model is trained on $D_{\textit{Wiki}}$ and tested on $D_{\textit{Blind}}$. \textsc{BASELINE} was similarly trained only on $D_{\textit{Wiki}}$ , making it a fair comparison. Table \ref{tab:holdOutExp} shows the results\footnote{For \textsc{BASELINE} \cite{deri2015make}, we use their trained model from \url{http://leps.isi.edu/fst/step-all.php} }. Our best model gets \textit{Distance} of $1.96$ as compared to $2.32$ from \textsc{BASELINE}.

\begin{figure}
\captionsetup[subfigure]{labelformat=empty}
\captionsetup[subfigure]{justification=centering}
\begin{subfigure}{0.2\textwidth}
\includegraphics[scale=0.20]{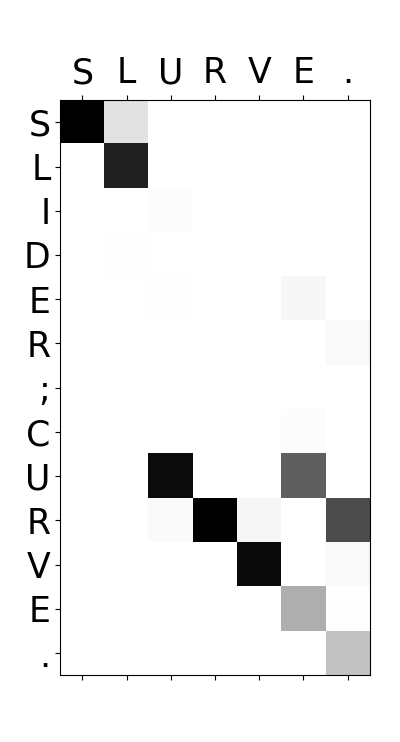}
\end{subfigure} 
\begin{subfigure}{0.2\textwidth}
\centering
\includegraphics[scale=0.21]{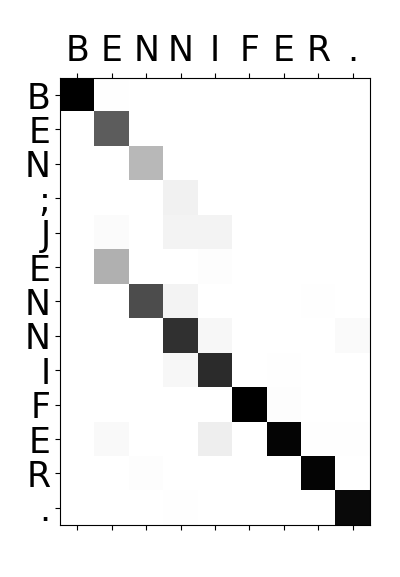}
\end{subfigure}
\caption{Attention matrices while generating \textit{slurve} from \textit{slider;curve}, and \textit{bennifer} from \textit{ben;jennifer} respectively, using \textit{Forward} model. \textbf{;} and \textbf{.} are  separator and stop characters. Darker cells are higher-valued}
\textbf{\label{fig:attention}}
\end{figure}

 We observe that the \emph{Backward} architecture 
 performs better than \emph{Forward} architecture, confirming our hypothesis in \S\ref{subsection:back}.
 In addition, ablation results confirm the importance of attention, and initializing the word embeddings.
 We believe this is because portmanteaus have high fidelity towards their root word characters and its critical that the model can observe all root sequence characters, which attention manages to do as shown in Fig.~\ref{fig:attention}.

\begin{table}
\centering
\scriptsize
\begin{tabular}{|l|l|l|l|l|l|l| }
\hline 
Model & Attn & Ens & Init & Search & Matches & Distance\\ \hline \hline
\textsc{Baseline} &  {-} & {-}  & {-} &  {-} & \textbf{31.56\%} & 2.32  \\ \hline
\multirow{4}{*}{\textsc{Forward}} & $\checkmark$ & $\times$ & $\checkmark$ & \textsc{SCORE} & 25.26\% & 2.13 \\ 
 & $\checkmark$ & $\times$ & $\times$ & \textsc{SCORE} & 24.93\% & 2.32 \\ 
 & $\checkmark$ & $\checkmark$ & $\times$ & \textsc{SCORE} & 31.23\% & 1.98 \\ 
 & $\checkmark$ & $\checkmark$ & $\checkmark$ & \textsc{SCORE} & 28.94\% & 2.04 \\ \hline
\multirow{4}{*}{\textsc{Backward}} & $\checkmark$ & $\times$ & $\checkmark$ & \textsc{SCORE} & 25.75\% & 2.14 \\ 
& $\checkmark$ & $\times$ & $\times$ & \textsc{SCORE} & 25.26\% & 2.17 \\ 
& $\checkmark$ & $\checkmark$ & $\times$ & \textsc{SCORE} & \textbf{31.72\%} & \textbf{1.96} \\ 
& $\checkmark$ & $\checkmark$ & $\checkmark$ & \textsc{SCORE} &  \textbf{32.78\%} & \textbf{1.96} \\ \hline
\end{tabular}
\caption{Results on $D_{\textit{Blind}}$ (1223 Examples). In general, \textsc{Backward} architecture performs better than \textsc{Forward} architecture.}
\textbf{\label{tab:holdOutExp}}
\end{table}

\subsubsection{Performance on Uncovered Examples}
The set of candidates generated before scoring in the approximate \textsc{SCORE} decoding approach sometimes do not cover the ground truth. This holds true for 229 out of 1223 examples in $D_{Blind}$. We compare the \textsc{Forward} approach along with a \textsc{Greedy} decoding strategy to the \textsc{Baseline} approach for these examples.

Both \textsc{Forward+Greedy} and the \textsc{Baseline} get 0 \textit{Matches} on these examples. The \textit{Distance} for these examples is 4.52 for \textsc{Baseline} and 4.09 for \textsc{Forward+Greedy}. Hence, we see that one of our approaches (\textsc{Forward+Greedy}) outperforms \textsc{Baseline} even for these  examples.

\subsection{Significance Tests}
Since our dataset is still small relatively small ($1223$ examples), it is essential to verify whether \textsc{Backward} is indeed statistically significantly better than \textsc{Baseline} in terms of \textit{Matches}. 

In order to do this, we use a paired bootstrap\footnote{We average across $M=1000$ randomly chosen subsets of $D_{Blind}$, each of size $N=611$ ($\approx 1223/2$)} comparison \cite{koehn2004statistical} between \textsc{Backward} and \textsc{Baseline} in terms of \textit{Matches}. \textsc{Backward} is found to be better (gets more \textit{Matches}) than \textsc{Baseline} in 99.9\% ($p=0.999$) of the subsets. 

Similarly, \textsc{Backward} has a lower \textit{Distance} than \textsc{Baseline} by a margin of $0.2$ in 99.5\% ($p=0.995$) of the subsets. 

\subsection{Subjective Evaluation and Analysis}

On inspecting outputs, we observed that often output from our system seemed good in spite of high edit distance from ground truth. Such aspect of an output \emph{seeming good} is not captured satisfactorily by measures like edit distance.
To compare the errors made by our model to the baseline, we designed and conducted a human evaluation task on AMT.%
\footnote{We avoid ground truth comparison because annotators can be biased to ground truth due to its existing popularity.}
In the survey, we show human annotators outputs from our system and that of the baseline.
We ask them to judge which alternative is \emph{better} overall based on following criteria:  1. It is a good shorthand for two original words  2. It sounds better. We requested annotation on a scale of 1-4. To avoid ordering bias, we shuffled the order of two portmanteau between our system and that of baseline. We restrict annotators to be from Anglophone countries, have HIT Approval Rate $>$ $80$\% and pay $0.40$\$ per HIT (5 Questions per HIT).

As seen in Table \ref{tab:XX}, output from our system was labelled better by humans as compared to the baseline 58.12\% of the time. Table \ref{table:examples} shows outputs from different models for a few examples.

\begin{table}
\centering
\scriptsize
\begin{tabular}{|c|c|c|c| }
\hline 
Input & \textsc{forward} & \textsc{backward} & \textsc{Ground Truth}  \\ \hline 
shopping;marathon & shopparathon & shoathon & shopathon \\
fashion;fascism & fashism & fashism & fashism \\
wiki;etiquette & wikiquette & wiquette & wikiquette  \\
clown;president & clowident & clownsident & clownsident  \\
\hline
\end{tabular}
\caption{Example outputs from different models (Refer to appendix for more examples)}
\textbf{\label{table:examples}}
\end{table}


\begin{table}
\centering
\scriptsize
\begin{tabular}{|c|c|}
\hline 
Judgement & Percentage of total  \\ \hline \hline
Much Better (1) & 29.06  \\ \hline 
Better (2) & 29.06 \\ \hline 
Worse (3) & 25.11 \\ \hline 
Much Worse (4) & 16.74 \\ \hline 
\end{tabular}
\caption{AMT annotator judgements on whether our system's proposed portmanteau is better or worse compared to the baseline}\textbf{\label{tab:XX}}
\end{table}

\section{Related Work}


\newcite{ozbal2012computational} generate new words to describe a product given its category and properties. However, their method is limited to hand-crafted rules as compared to our data driven approach. Also, their focus is on brand names. \newcite{namification} have proposed an approach to recommend brand names based on brand/product description. However, they consider only a limited number of features like memorability and readability. 
\newcite{smith2014nehovah} devise an approach to generate portmanteaus, which requires user-defined weights for attributes like \textit{sounding good}. Generating a portmanteau from two root words can be viewed as a S2S problem. Recently, neural approaches  have been used for S2S problems \cite{sutskever2014sequence} such as MT. \newcite{ling2015character} and \newcite{chung2016character} have shown that character-level neural sequence models work as well as word-level ones for language modelling and MT. \newcite{zoph2016multi} propose S2S models for multi-source MT, which have multi-sequence inputs, similar to our case. 

\section{Conclusion}
We have proposed an end-to-end neural system to model portmanteau generation. Our experiments show the efficacy of proposed system in predicting portmanteaus given the root words. 
We conclude that pre-training character embeddings on the English vocabulary helps the model. Through human evaluation we show that our model's predictions are superior to the baseline. We have also released our dataset and code\footnote{\url{https://github.com/vgtomahawk/Charmanteau-CamReady}} to encourage further research on the phenomenon of portmanteaus. We also release an online demo \footnote{\url{http://tinyurl.com/y9x6mvy}} where our trained model can be queried for portmanteau suggestions. An obvious extension to our work is to try similar models on multiple languages.

\section*{Acknowledgements}
We thank Dongyeop Kang, David Mortensen, Qinlan Shen and anonymous reviewers for their valuable comments. This research was supported in part by DARPA grant FA8750-12-2-0342 funded under the DEFT program.





\bibliography{emnlp2017}

\begin{thebibliography}{}
\expandafter\ifx\csname natexlab\endcsname\relax\def\natexlab#1{#1}\fi

\bibitem[{Algeo(1977)}]{algeo1977blends}
John Algeo. 1977.
\newblock Blends, a structural and systemic view.
\newblock {\em American speech\/} 52(1/2):47--64.

\bibitem[{Bahdanau et~al.(2014)Bahdanau, Cho, and Bengio}]{bahdanau2014neural}
Dzmitry Bahdanau, Kyunghyun Cho, and Yoshua Bengio. 2014.
\newblock Neural machine translation by jointly learning to align and
  translate.
\newblock {\em arXiv:1409.0473\/} .

\bibitem[{Bat-El(1996)}]{bat1996selecting}
Outi Bat-El. 1996.
\newblock Selecting the best of the worst: the grammar of {Hebrew} blends.
\newblock {\em Phonology\/} 13(03):283--328.

\bibitem[{Berman(1989)}]{berman1989role}
Ruth Berman. 1989.
\newblock The role of blends in {Modern} {Hebrew} word-formation.
\newblock {\em Studia linguistica et orientalia memoriae Haim Blanc dedicata.
  Wiesbaden: Harrassowitz\/} pages 45--61.

\bibitem[{Breiman(1996)}]{breiman1996bagging}
Leo Breiman. 1996.
\newblock {Bagging} predictors.
\newblock {\em Machine learning\/} 24(2):123--140.

\bibitem[{Chung et~al.(2016)Chung, Cho, and Bengio}]{chung2016character}
Junyoung Chung, Kyunghyun Cho, and Yoshua Bengio. 2016.
\newblock A character-level decoder without explicit segmentation for neural
  machine translation.
\newblock {\em arXiv:1603.06147\/} .

\bibitem[{Dardjowidjojo(1979)}]{dardjowidjojo1979acronymic}
Soenjono Dardjowidjojo. 1979.
\newblock Acronymic {Patterns} in {Indonesian}.
\newblock {\em Pacific Linguistics Series C\/} 45:143--160.

\bibitem[{Deri and Knight(2015)}]{deri2015make}
Aliya Deri and Kevin Knight. 2015.
\newblock How to make a frenemy: Multitape {FSTs} for portmanteau generation.
\newblock In {\em Proceedings of NAACL-HLT\/}. pages 206--210.

\bibitem[{Faruqui et~al.(2016)Faruqui, Tsvetkov, Neubig, and
  Dyer}]{faruqui2016morphological}
Manaal Faruqui, Yulia Tsvetkov, Graham Neubig, and Chris Dyer. 2016.
\newblock Morphological {Inflection} {Generation} using {Character} {Sequence}
  to {Sequence} {Learning}.
\newblock In {\em Proceedings of NAACL-HLT\/}. pages 634--643.

\bibitem[{Gabler(2015)}]{nyt}
Neal Gabler. 2015.
\newblock The {Weird} {Science} of {Naming} {New} {Products}.
\newblock {\em New York Times - {\url{http://tinyurl.com/lmlq7ex}}\/} .

\bibitem[{Hiranandani et~al.(2017)Hiranandani, Maneriker, and
  Jhamtani}]{namification}
Gaurush Hiranandani, Pranav Maneriker, and Harsh Jhamtani. 2017.
\newblock Generating appealing brand names.
\newblock {\em arXiv preprint arXiv:1706.09335\/} .

\bibitem[{Hoang et~al.(2017)Hoang, Haffari, and Cohn}]{hoang2017decoding}
Cong Duy~Vu Hoang, Gholamreza Haffari, and Trevor Cohn. 2017.
\newblock {Decoding} as {Continuous} {Optimization} in {Neural} {Machine}
  {Translation}.
\newblock {\em arXiv:1701.02854\/} .

\bibitem[{Koehn(2004)}]{koehn2004statistical}
Philipp Koehn. 2004.
\newblock Statistical significance tests for machine translation evaluation.
\newblock In {\em EMNLP\/}. pages 388--395.

\bibitem[{Ling et~al.(2015)Ling, Trancoso, Dyer, and Black}]{ling2015character}
Wang Ling, Isabel Trancoso, Chris Dyer, and Alan~W Black. 2015.
\newblock Character-based neural machine translation.
\newblock {\em arXiv:1511.04586\/} .

\bibitem[{{\"O}zbal and Strapparava(2012)}]{ozbal2012computational}
G{\"o}zde {\"O}zbal and Carlo Strapparava. 2012.
\newblock A computational approach to the automation of creative naming.
\newblock In {\em Proceedings of ACL\/}. Association for Computational
  Linguistics, pages 703--711.

\bibitem[{Petri(2012)}]{washingtonpost}
Alexandra Petri. 2012.
\newblock {Say} {No} to {Portmanteaus}.
\newblock {\em Washington Post - {\url{http://tinyurl.com/kvmep2t}}\/} .

\bibitem[{Pi{\~n}eros(2004)}]{pineros2004creation}
Carlos-Eduardo Pi{\~n}eros. 2004.
\newblock The creation of portmanteaus in the extragrammatical morphology of
  {Spanish}.
\newblock {\em Probus\/} 16(2):203--240.

\bibitem[{Shaw et~al.(2014)Shaw, White, Moreton, and
  Monrose}]{shaw2014emergent}
Katherine~E Shaw, Andrew~M White, Elliott Moreton, and Fabian Monrose. 2014.
\newblock {Emergent} faithfulness to morphological and semantic heads in
  lexical blends.
\newblock In {\em Proceedings of the Annual Meetings on Phonology\/}. volume~1.

\bibitem[{Smith et~al.(2014)Smith, Hintze, and Ventura}]{smith2014nehovah}
Michael~R Smith, Ryan~S Hintze, and Dan Ventura. 2014.
\newblock Nehovah: A neologism creator nomen ipsum.
\newblock In {\em Proceedings of the International Conference on Computational
  Creativity\/}. pages 173--181.

\bibitem[{Sutskever et~al.(2014)Sutskever, Vinyals, and
  Le}]{sutskever2014sequence}
Ilya Sutskever, Oriol Vinyals, and Quoc~V Le. 2014.
\newblock Sequence to sequence learning with neural networks.
\newblock In {\em Neural information processing systems\/}. pages 3104--3112.

\bibitem[{Weide(1998)}]{weide1998cmu}
R~Weide. 1998.
\newblock The {CMU} pronunciation dictionary, release 0.6.
\newblock {\em Carnegie Mellon University\/} .

\bibitem[{Yu et~al.(2016)Yu, Blunsom, Dyer, Grefenstette, and
  Kocisky}]{yu2016neural}
Lei Yu, Phil Blunsom, Chris Dyer, Edward Grefenstette, and Tomas Kocisky. 2016.
\newblock The {Neural} {Noisy} {Channel}.
\newblock {\em arXiv:1611.02554\/} .

\bibitem[{Zoph and Knight(2016)}]{zoph2016multi}
Barret Zoph and Kevin Knight. 2016.
\newblock Multi-source neural translation.
\newblock {\em arXiv:1601.00710\/} .

\end{thebibliography}
\bibliographystyle{emnlp_natbib}

\end{document}